\def\year{2021}\relax
\documentclass[letterpaper]{article} 
\usepackage{aaai21}  
\usepackage{times}  
\usepackage{helvet} 
\usepackage{courier}  
\usepackage[hyphens]{url}  
\usepackage{graphicx} 
\usepackage{natbib}  
\usepackage{caption} 
\usepackage{float}
\usepackage{subcaption}
\usepackage{booktabs}
\usepackage{hyperref}
\usepackage{xcolor}
\usepackage{lipsum}  
\usepackage{array}
\usepackage{graphicx}
\usepackage{multirow}

\newcommand\MyBox[2]{
  \fbox{\lower0.75cm
    \vbox to 1.7cm{\vfil
      \hbox to 1.7cm{\hfil\parbox{1.4cm}{#1\\#2}\hfil}
      \vfil}%
  }%
}
\definecolor{bleudefrance}{rgb}{0.19, 0.55, 0.91}
\definecolor{ceruleanblue}{rgb}{0.16, 0.32, 0.75}
\definecolor{mediumblue}{rgb}{0.0, 0.0, 0.8}
\definecolor{grey}{HTML}{969696}
\definecolor{violet}{HTML}{756bb1}

\usepackage[acronym,nonumberlist]{glossaries}
\newacronym{dl}{DL}{Deep Learning}
\newacronym{gpu}{GPU}{Graphics Processing Unit}
\newacronym{cpu}{CPU}{Central Processing Unit}
\newacronym{tpu}{TPU}{Tensor Processing Unit}
\newacronym{ecmwf}{ECMWF}{European Centre for Medium-Range Weather Forecasts}
\newacronym{imerg}{IMERG}{Integrated Multi-satellitE Retrievals}
\newacronym{simsat}{SimSat}{simulated satellite data}
\newacronym{ml}{ML}{Machine Learning}
\newacronym{rmse}{RMSE}{Root-Mean-Square Error}

\usepackage[binary-units]{siunitx}
\usepackage{cleveref}
\crefname{table}{Table}{Tables}
\crefname{figure}{Figure}{Figures}
\crefname{section}{Section}{Sections}

\title{RainBench: Towards Data-Driven Global Precipitation Forecasting\\from Satellite Imagery}
\author{
Anonymous Authors
}

\title{RainBench: Towards Global Precipitation Forecasting from Satellite Imagery}
\author {
        Christian Schroeder de Witt*\textsuperscript{\rm 1},
        Catherine Tong*\textsuperscript{\rm 1},
        Valentina Zantedeschi\textsuperscript{\rm 2},
        Daniele De Martini\textsuperscript{\rm 1},
        Freddie Kalaitzis\textsuperscript{\rm 1},
        Matthew Chantry\textsuperscript{\rm 1},
        Duncan Watson-Parris\textsuperscript{\rm 1},
        Piotr Biliński\textsuperscript{\rm 3}\\
}
\affiliations {
    \textsuperscript{\rm 1} University of Oxford (*: Equal contribution)\\
    \textsuperscript{\rm 2} Inria, Lille - Nord Europe research centre and University College London, Centre for Artificial Intelligence \\
    \textsuperscript{\rm 3} University of Warsaw\\
}
\begin{document}
\maketitle

\begin{abstract}

Extreme precipitation events, such as violent rainfall and hail storms, routinely ravage economies and livelihoods around the developing world. Climate change further aggravates this issue \cite{gupta2020effects}. Data-driven deep learning approaches could widen the access to accurate multi-day forecasts, to mitigate against such events. However, there is currently no benchmark dataset dedicated to the study of global precipitation forecasts. In this paper, we introduce \textbf{RainBench}, a new multi-modal benchmark dataset for data-driven precipitation forecasting. It includes simulated satellite data, a selection of relevant meteorological data from the ERA5 reanalysis product, and IMERG precipitation data. We also release \textbf{PyRain}, a library to process large precipitation datasets efficiently. We present an extensive analysis of our novel dataset and establish baseline results for two benchmark medium-range precipitation forecasting tasks. Finally, we discuss existing data-driven weather forecasting methodologies and suggest future research avenues.


\end{abstract}


\glsresetall
\glsunset{cpu}
\glsunset{gpu}

\section{Introduction}

Extreme precipitation events, such as violent rain and hail storms, can devastate crop fields and disrupt harvests \cite{vogel2019effects,li_excessive_2019}.
These events can be locally forecasted
with sophisticated numerical weather models that rely on extensive ground and satellite observations. However, such approaches require access to compute and data resources that developing countries in need - particularly in South America and West Africa - cannot afford
\cite{le2020comparison,gubler2020assessment}.
The lack of advance planning for precipitation events impedes socioeconomic development and ultimately affects the livelihoods of millions around the world.
Given the increase in global precipitation and extreme precipitation events driven by climate change \cite{gupta2020effects}, the need for accurate precipitation forecasts is ever more pressing.

Data-driven machine learning approaches circumvent the dependence on traditional resource-intensive numerical models, which typically
take several hours to run \cite{metnet}, incurring a significant time lag.
In contrast, deep learning models deployed on dedicated high-throughput hardware can produce inferences in a matter of seconds. However, while there have been attempts in forecasting precipitation with neural networks, they have mostly been fragmented across different local regions, which hinders a systematic comparison into their performance.

In this work, we introduce \textbf{RainBench}, a multi-modal dataset to support data-driven forecasting of global precipitation from satellite imagery. We curate three types of datasets: simulated satellite data (SimSat), numerical reanalysis data (ERA5), and global precipitation estimates (IMERG).
The use of satellite images to forecast precipitation globally would circumvent the need to collect ground station data, and hence they are key to our vision for widening the access to multi-day precipitation forecasts.
Reanalysis data provide estimates of complete atmospheric state, and IMERG provides rigorous estimates of global precipitation. Access to these data opens up opportunities to develop more timely and potentially physics-informed forecast models, which so far could not have been studied systematically.

Most related to our work,
\citet{rasp_weatherbench_2020}
have developed WeatherBench, a benchmark environment for global data-driven medium-range weather forecasting. This dataset forms an excellent first step in weather forecasting.
However, some important features of WeatherBench limit its use for end-to-end precipitation forecasts.
WeatherBench does not include any observational raw data (e.g. satellite data) and only contains ERA5 reanalysis data, which have limited resolution of extreme precipitation events. Further, WeatherBench does not include a fast dataloading pipeline to train ML models, which we found to be a significant bottleneck in our model development and testing process.
This gap prompted us to also
release \textbf{PyRain}, a data processing and experimentation framework with fast and configurable multi-modal dataloaders.

To summarise our contributions:
(a) We introduce the multi-modal \textbf{RainBench} dataset which supports data-driven investigations for global precipitation forecasting from satellite imagery;
(b) we release \textbf{PyRain}, which allows researchers to run \gls{dl} experiments on RainBench efficiently, reducing time and hardware costs and thus lowering the barrier to entry into this field;
(c) we introduce two benchmark precipitation forecasting tasks on RainBench and their baseline results, and present experiments studying class-balancing schemes.
Finally, we discuss the challenges in the field and outline several fruitful avenues for future research.

\section{Related Work}
\label{sec:rel}

Weather forecasting systems have not fundamentally changed since they were first operationalised nearly 50 years ago. Current state-of-the-art operational weather forecasting systems rely on numerical models that forward the physical atmospheric state in time based on a system of physical equations and parameterised subgrid processes \cite{bauer2015quiet}. While global simulations typically run at grid sizes of \SI{10}{\kilo\metre}, regional models can reach \SI{1.5}{\kilo\metre} \cite{franch_taasrad19_2020} .
Even in the latter case, skilled forecast lengths are usually limited to a maximum of \num{10} days, with a conjectured hard limit of \num{14} to \num{15} days \cite{zhang_what_2019}.
\textit{Nowcasting}, i.e. high-resolution weather forecasting only a few hours in advance, is currently limited by the several hours that numerical forecasting models take to run \cite{metnet}.

Given the huge amounts of data currently available from both numerical models and observations, new opportunities exist to train data-driven models to produce these forecasts. The current boom in \gls{ml} has inspired several other groups to approach the problem of weather forecasting. Early work by \citeauthor{xingjian2015convolutional} have invested using convolutional recurrent neural networks for precipitation nowcasting. More recently, \citeauthor{metnet} from Google proposed a ``(weather) model free" approach, MetNet, which seeks to forecast precipitation in continental USA using geostationary satellite images and radar measurements as inputs. This approach performs well up to 7-8 hours, but inevitably runs into a forecast horizon limit as information from global or surrounding geographic areas is not incorporated into the system. This time window has value though it would not enable substantial disaster preparedness.

The prediction of extreme precipitation (and other extreme weather events) has a long history with traditional forecasting systems \cite{lalaurette_early_2003}. More recent developments in ensemble weather forecasting systems surround the introduction of novel forecasting indices \cite[EFI]{zsoter_recent_2006} and post-processing \cite{gronquist_deep_2020}. There has also been other deep-learning based precipitation forecasting models as motivated by the monsoon prediction problem, for example, \citet{saha_deep_2017} and \citet{saha_prediction_2020}
use a stacked autoencoder to identify climatic predictors and an ensemble regression tree model, while \citet{praveen_analyzing_2020} use kriging and multi-layer perceptrons to predict monsoon rainfall from ERA5 data.   

WeatherBench \cite{rasp_weatherbench_2020} is a benchmark dataset for data-driven global weather forecasting, derived from data in the ERA5 archive. Its release has prompted a number of follow-up works to employ deep learning techniques for weather forecasting, although the variables considered have only been restricted to the forecasts of relatively static variables, such as 500 hPa geopotential and 850 hPa temperature \cite{weyn_can_2019,weyn_improving_2020,rasp2020purely,bihlo2020generative, arcomano2020machine}. Unlike RainBench which incorporates the element of observational input data from (simulated) satellites, WeatherBench's data comes solely from the ERA5 reanalysis archive, and thus provides no route to producing an end-to-end forecasting system.







\section{RainBench}
\label{sec:extreme}

In this section, we introduce RainBench, which consists of data derived from three publicly-available sources: (1) \gls{ecmwf} \gls{simsat}, (2) the ERA5 reanalysis product, and (3) \gls{imerg} global precipitation estimates. 


\vspace{-1em}
\paragraph{SimSat} We use simulated satellite data in place of real satellite imagery to minimise data processing requirements and to simplify the prediction task. \Gls{simsat} data are model-simulated satellite data generated from \gls{ecmwf}'s high-resolution weather-forecasting model using the RTTOV radiative transfer model \cite{rttov}. 
\Gls{simsat} emulates three spectral channels from the Meteosat-10 SEVIRI satellite \cite{aminou2002msg}.
\Gls{simsat} provides information about global cloud cover and moisture features and has a native spatial resolution of about \SI{0.1}{\degree} -- i.e. about \SI{10}{\kilo\metre} -- at three-hourly intervals. The product is available from April 2016 to present (with a lag time of \SI{24}{\hour}).
Using simulated satellite data provides an intermediate step to using real satellite observations as the images are a global nadir view of Earth, avoiding issues of instrument error and large numbers of missing values.
Here we aggregate the data to \SI{0.25}{\degree} -- about \SI{30}{\kilo\metre} -- to be consistent with the ERA5 dataset.

\vspace{-1em}
\paragraph{ERA5} We use ERA5 as it is an accurate and commonly used reanalysis product familiar to the climate science community \cite{rasp_weatherbench_2020}. ERA5 reanalysis data provides hourly estimates of a variety of atmospheric, land and oceanic variables, such as specific humidity, temperature and geopotential height at different pressure levels \cite{era5}.
Estimates cover the full globe at a spatial resolution of \SI{0.25}{\degree} and are available from 1979 to present, with a lag time of five days. 

\vspace{-1em}
\paragraph{IMERG} \Gls{imerg} is a global half-hourly precipitation estimation product provided by NASA \cite{imerg}. We use the Final Run product which primarily uses satellite data from multiple polar-orbiting and geo-stationary satellites. This estimate is then corrected using data from reanalysis products (MERRA2, ERA5) and rain-gauge data.
\Gls{imerg} is produced at a spatial resolution of \SI{0.1}{\degree} -- about \SI{10}{\kilo\metre} -- and is available from June 2000 to present, with a lag time of about three to four months.

To facilitate efficient experimentation, all data is converted from thier original resolutions to \SI{5.625}{\degree} resolutions using bilinear interpolation. 
 
RainBench provides precipitation values from two sources, ERA5 and IMERG, as both are widely used and considered to be high-quality precipitation datasets. The ERA5 precipitation is accumulated precipitation over the last hour and is calculated as an averaged quantity over a grid-box. We aggregated \Gls{imerg} precipitation into hourly accumulated precipitation and should be considered as a point estimate of the precipitation.





\begin{figure}[h]
\includegraphics[width=8cm]{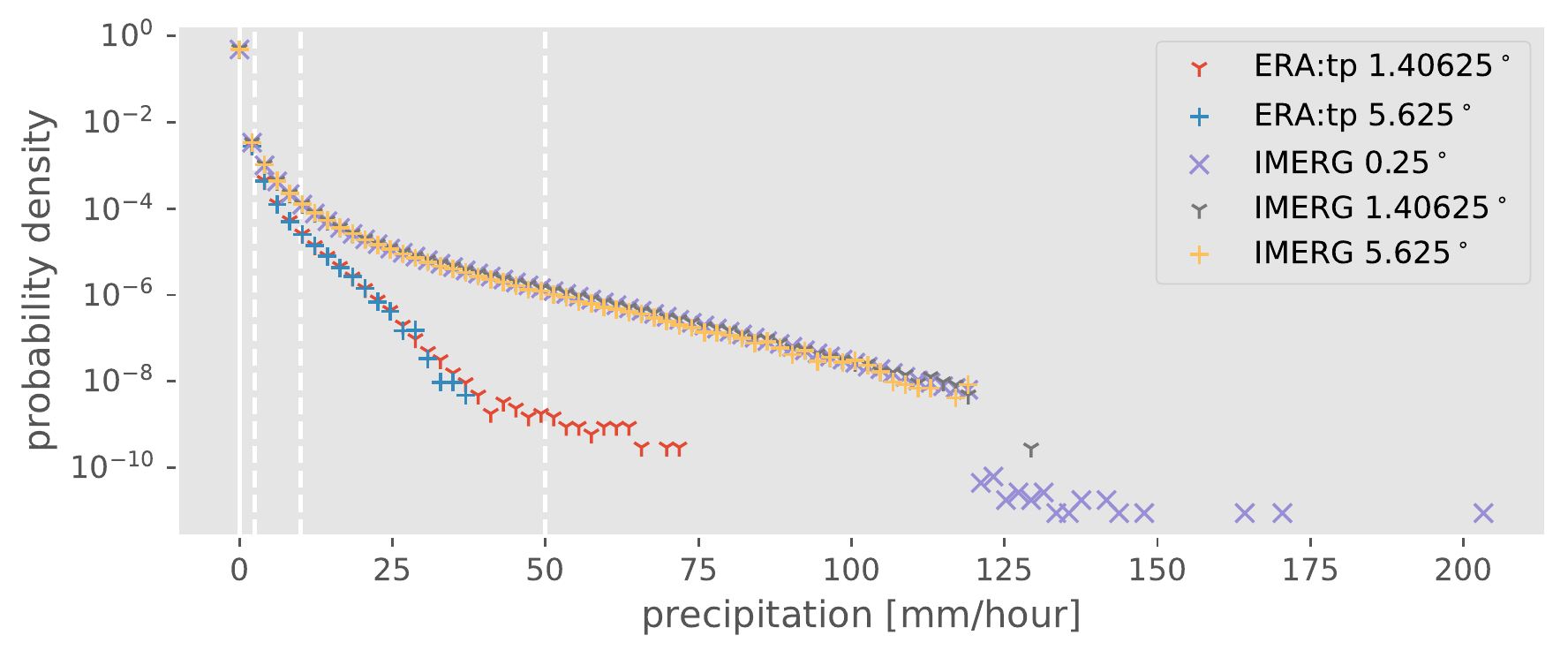}
\caption{Precipitation histogram from 2000-2017 with ERA5 and \gls{imerg} at different resolutions. Vertical lines delineate convection rainfall types: slight (\SIrange{0}{2}{\milli\meter\per\hour}), moderate (\SIrange{2}{10}{\milli\meter\per\hour}), heavy (\SIrange{10}{50}{\milli\meter\per\hour}), and violent (over \SI{50}{\milli\meter\per\hour}) \cite{metoffice_fact_2012}.}
\label{fig:hist_rain}
\end{figure}

 \autoref{fig:hist_rain} shows the distribution of precipitation for the years 2000-2017 with both ERA5 and \gls{imerg}. \gls{imerg} is generally regarded as a more trust-worthy dataset for precipitation due to the direct inclusion of precipitation observations in the data assimilation process and the higher spatial resolution used to produce the dataset, which also result in seen difference in data distributions. \gls{imerg} has significantly larger rainfall tails than ERA5, and these tails rapidly vanish with decreasing dataset resolution. The underestimation of extreme precipitation events in ERA5 is clearly visible.
 
 

\section{PyRain}
To support efficient data-handling and experimentation on Rainbench, we release PyRain, an out-of-the-box experimentation framework. 


PyRain\footnote{https://github.com/frontierdevelopmentlab/pyrain} introduces an efficient dataloading pipeline for complex sample access patterns that scales to the terabytes of spatial timeseries data typically encountered in the climate and weather domain.
Previously identified as a decisive bottleneck by the Pangeo community\footnote{\url{https://pangeo.io/index.html} (2021)}, PyRain overcomes existing dataloading performance limitations through an efficient use of NumPy \textit{memmap} arrays\footnote{\url{https://docs.python.org/3/library/mmap.html} (2021)} in conjunction with optimised software-side access patterns.

In contrast to storage formats requiring \textit{read} system calls, including HDF5\footnote{\url{https://portal.hdfgroup.org/display/HDF5/HDF5}(2021)}, Zarr\footnote{\url{https://zarr.readthedocs.io/en/stable/} (2021)} or xarray\footnote{\url{http://xarray.pydata.org/en/stable/} (2021)}, memory-mapped files use the \textit{mmap} system call to map physical disk space directly to virtual process memory, enabling the use of \textit{lazy} OS demand paging and circumventing the kernel buffer. 
While less beneficial for chunked or sequential reads and spatial slicing, memmaps can efficiently handle the fragmented random access inherent to the randomized sliding-window access patterns along the primary axis as required in model training.  

In \autoref{tab:speedup}, we compare PyRain's memmap data reading capcity against a NetCDF+Dask \footnote{\url{https://www.unidata.ucar.edu/software/netcdf/} (2021)}  \cite{dask} dataloader. We find empirically that PyRain's memmap dataloader offers significant speedups over other solutions, saturating even SSD I/O with few process workers when used with PyTorch's \cite{pytorch} inbuilt dataloader.



Note that explicitly storing each training sample is not only slow and inflexible for research settings, but it also requires twenty to fifty times more storage and as a result comes at a higher cost than constructing samples on-the-fly. Thus, other options such as writing samples in TFRecord format \cite{weyn_can_2019,abadi_tensorflow_2016} would only be sensible for highly distributed training in production settings.

PyRain's dataloader is easily configurable and supports both complex multimodal item compositions, as well as periodic \cite{metnet} and sequential \cite{weyn_improving_2020} train-test set partitionings.
Apart from its data-loading pipeline, PyRain also supplies flexible raw-data conversion tools, a convenient interface for data-analysis tasks, various data-normalisation methods and a number of ready-built training settings based on PyTorch Lightning\footnote{\url{https://pytorch-lightning.readthedocs.io/en/latest/} (2021)}.
While being optimised for use with RainBench, PyRain is also compatible with WeatherBench.




\begin{table}[t]
\begin{center}
\caption{Number of data samples loaded per second using PyRain versus a conventional NetCDF framework. Typical configurations assumed and performed on a NVIDIA DGX1 server with $64$ CPUs.}
\label{tab:speedup}
\begin{tabular}{@{\extracolsep{5pt}} lccc}
      \toprule
  & NetCDF & PyRain & Speedup \\ 
 \midrule
16 workers & 40 & 2410 & 60.3$\times$ \\
64 workers & 70 & 1930 & 27.6$\times$ \\
\bottomrule
\end{tabular}
\end{center}
\end{table}

\section{Evaluation Tasks} \label{sec:tasks}



We define two benchmark tasks on RainBench for precipitation forecasting, with the ground truth precipitation values taken from either ERA5 or \gls{imerg}.


For each benchmark task, we consider three different input data settings: \gls{simsat}, reanalysis data (ERA5), or both. From the ERA5 dataset, we select a subset of variables as input to the forecast model based on our data analysis results; the inputs are geopotential (z), temperature (t), humidity (q), cloud liquid water content (clwc), cloud ice water content (ciwc), each sampled at \SIlist{300;500;850}{\hecto\pascal} geopotential heights; to these we add the surface pressure and the 2-meter temperature (t2m), as well as static variables that describe the location and surface of the Earth, i.e. latitude, longitude, land-sea mask, orography and soil type.
From the \gls{simsat} dataset, the inputs are cloud-brightness temperature (clbt) taken at three wavelengths. We normalize each variable with its global mean and standard deviation.

Since data from each source are available at different times, we use the data subset from April 2016 to train all models for the benchmark tasks, unless specified otherwise. We use data from 2018 and 2019 as validation and test sets respectively. To make sure no overlap exists between training and evaluation data, the first evaluated date is 6 January 2019 while the last training date is 31 December 2017. 




\begin{figure}[t!]
    \centering
    \includegraphics[width=8cm]{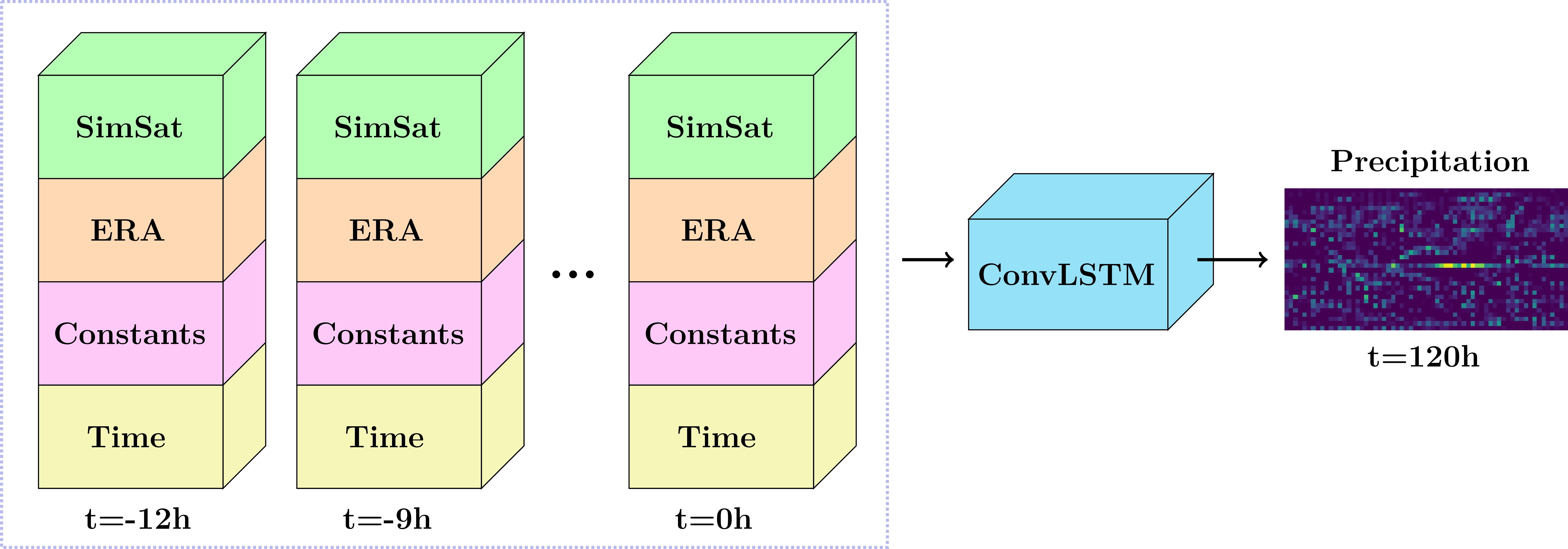}
    \caption{Model setup for the benchmark forecasting tasks.}
    \label{fig:approach}
\end{figure}

We perform experiments with a neural network based on Convolutional LSTMs, which have been shown to be effective for regional precipitation nowcasting \cite{xingjian2015convolutional}.
We structure our forecasting task based on MetNet's configurations~\cite{metnet}, where a single model is trained conditioned on time and is capable of forecasting at different lead times.

The network's input is composed of a time series $\{x_{t}\}$, where each $x_t$ is the set of standardized features at time $t$, sampled in regular intervals $\Delta t$ from $t=-T$ to $t=0$; the output is a precipitation forecast $y$ at lead time $t=\tau \leq {\tau}_L$.
In addition to the aforementioned atmospheric features, static features (e.g. latitude) along with three time-dependant features (hour, day, month) are repeated per timestep.
The input vector is then concatenated with a lead-time one-hot vector $x_{\tau}$. In our experiments, we adopt $T=12$ \si{\hour}, $\Delta t = 3$ \si{\hour} and forecasts at 24-hour intervals up to $\tau_{L} = 120$ \si{\hour}.
We note that we do not include precipitation as an input temporal feature. An overview of our setup is shown in \autoref{fig:approach}.

We approach the tasks as a regression problem. Following \cite{rasp_weatherbench_2020}, we use the mean latitude-weighted \gls{rmse} as loss and evaluation metric. We compare the results to persistence and climatology baselines. For persistence, precipitation values at $t=0$ are used as prediction at $t=\tau$. We compute  climatology and weekly climatology baselines from the full training dataset (since 1979 for ERA5 and since 2000 for IMERG), where local climatologies are computed as a single mean over all times and per week respectively \cite{rasp_weatherbench_2020}.

\section{Results}
\label{sec:results}

In this section, we first present our data analysis of RainBench. We then describe models' performance on the benchmark precipitation forecasting tasks, which highlights the difficulty in forecasting precipitation values on IMERG. Finally, we present an experiment on same-timestep precipitation estimation to investigate class balancing issues.

\subsection{Data Analysis}

\begin{figure}[t]
\includegraphics[width=8.5cm]{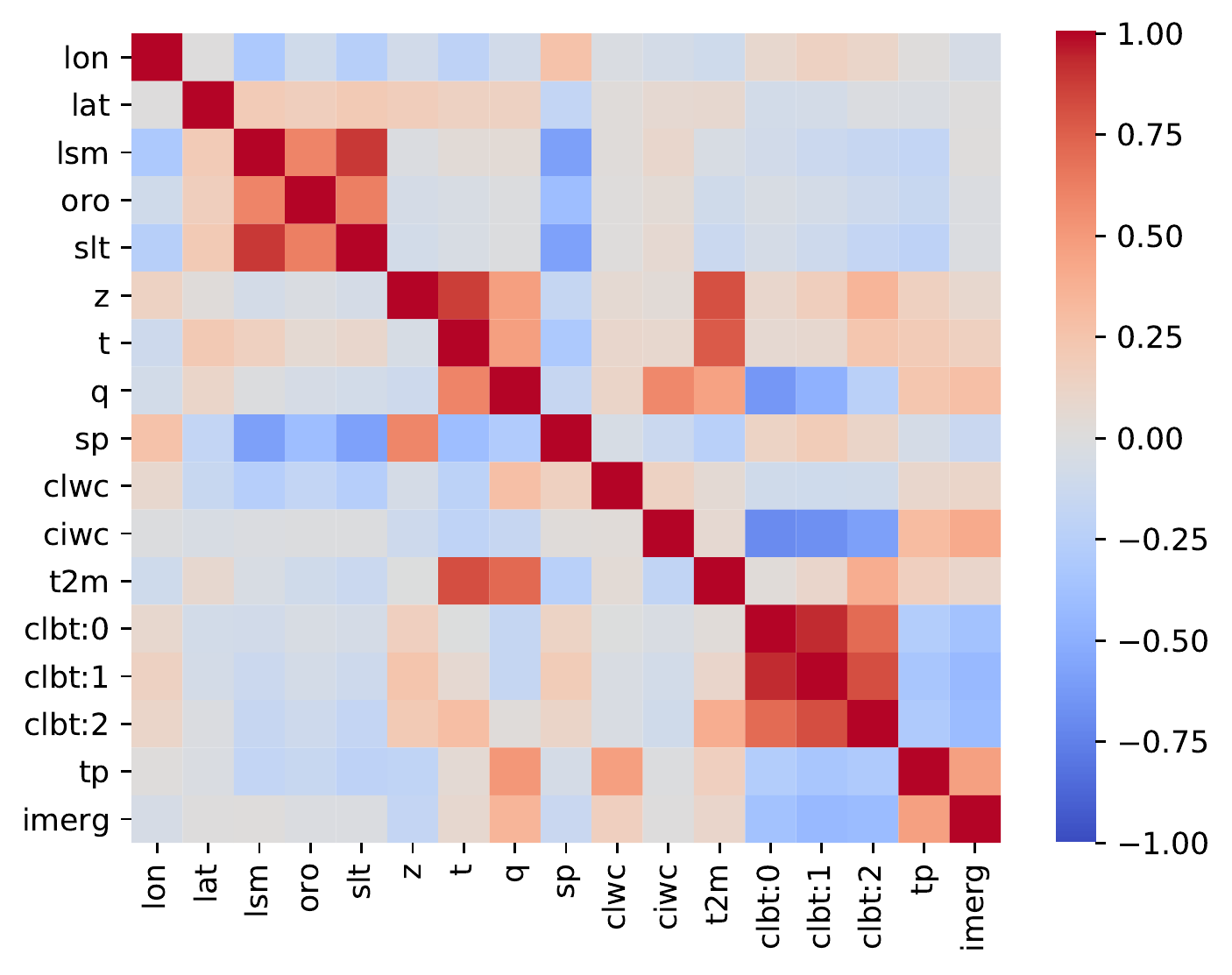}
\caption{Spearman's correlation of RainBench variables from April 2016 to December 2019 in latitude band $[-60^\circ,60^\circ]$ at pressure levels \SI{300}{\hecto\pascal} (about \SI{10}{\kilo\metre}) (upper triangle) and \SI{850}{\hecto\pascal} (\SI{1.5}{\kilo\metre}) (lower triangle). Legend: $lon$: longitude, $lat$: latitude, $lsm$: land-sea mask, $oro$: orography (topographic relief of mountains), $lst$: soil type, $z$: geopotential height, $t$: temperature, $q$: specific humidity, $sp$: surface pressure, $\text{clwc}$: cloud liquid water content, $\text{ciwc}$: cloud ice water content, $t2m$: temperature at 2m, $\text{clbt:}i$: $i$th SimSat channel, $tp$: ERA5 total precipitation, $\text{imerg}$: IMERG precipitation. All correlations in this plot are statistically significant ($p<0.05$).}
\label{fig:correlation}
\end{figure}

To analyse the dependencies between all RainBench variables, we calculate pairwise Spearman's rank correlation indices over latitude band from \num{-60} to \SI{60}{\degree} and date range from April 2016 to December 2019 (see \autoref{fig:correlation}). 
In contrast to Pearson's correlation coefficient, Spearman's correlation coefficient is significant if there is a, potentially non-linear, monotonic relationship between variables, while Pearson's considers only linear correlations. This allows to capture relationships between variables such as between temperature and absolute latitude. 
Comparing correlations at altitude pressure levels \SI{300}{\hecto\pascal} (about \SI{10}{\kilo\metre}) and \SI{850}{\hecto\pascal} (\SI{1.5}{\kilo\metre}), we can see that they are almost identical, save for a few exceptions:
Specific humidity, $q$, and geopotential height, $z$, correlate strongly at \SI{300}{\hecto\pascal} but not at \SI{850}{\hecto\pascal}, cloud ice water content, $\text{ciwc}$, generally correlates more strongly at higher altitude (and cloud liquid water content, $\text{clwc}$, vice versa). A careful examination of the underlying physical dependencies results in the realisation that all of these asymmetries stem mostly from latitudinal correlations or effects related to cloud formation, e.g. ice and liquid form in clouds at different temperatures/altitudes. 

As we are particularly interested in variables that have predictive skill on precipitation, we note that all SimSat spectral channels moderately anti-correlate with both ERA5 and IMERG precipitation estimates. Interestingly, SimSat signals correlate much more strongly with specific humidity and cloud ice water content at higher altitude, which might be a consequence of spectral penetration depth. ERA5 state variables that correlate the most with either precipitation estimates are specific humidity and temperature. Cloud ice water content correlates moderately strongly with precipitation estimates at high altitude, but not at all at lower altitudes (where ice water content tends to be much lower). Further, a number of time-varying ERA5 state variables correlate more strongly with IMERG precipitation than ERA5 precipitation, as do SimSat signals. Conversely, a number of constant variables, such as land-sea mask, orography and soil type are significantly anti-correlated with ERA5 precipitation, but not at all correlated with IMERG. Overall, we find that all variables that are significantly correlated or anti-correlated with both ERA5 \text{tp} and IMERG are also correlated or anti-correlated with SimSat \text{clbt:0-2}, suggesting that precipitation prediction from simulated satellite data alone may be feasible.  


\subsection{Precipitation Forecasting}
\label{sec:fore}

\begin{table}[t]
  \centering
  \caption{Precipitation forecasts evaluated with Latitude-weighted RMSE (mm). All rows except where otherwise stated show models trained with data from 2016 onwards.
  }
  \label{tab:forecasts}
  \begin{subtable}[t]{\linewidth}
    \centering
    \vspace{0pt}
        \caption{Predicting Precipitation from ERA}
        \label{tab:forecasts_era}
    \begin{tabular}{@{\extracolsep{5pt}} lccc}
      \toprule
    Inputs & 1-day & 3-day & 5-day \\ 
      \midrule
    Persistence & 0.6249 &0.6460 & 0.6492 \\
    Climatology & \multicolumn{3}{c}{0.4492 (1979-2017)}  \\
    Climatology (weekly) & \multicolumn{3}{c}{\textbf{0.4447} (1979-2017)} \\
    \midrule
    SimSat & 0.4610 & 0.4678 & 0.4691 \\
    ERA & 0.4562 & 0.4655 & 0.4677  \\
    SimSat + ERA & \textbf{0.4557} & \textbf{0.4655} & \textbf{0.4675} \\
    \midrule
    ERA (1979-2017) & 0.4485 & 0.4670 & 0.4699 \\
      \bottomrule
    \end{tabular}
  \end{subtable}
  \begin{subtable}[t]{\linewidth}
    \centering
    \vspace*{0.3cm}
         \caption{Predicting Precipitation from IMERG}
         \label{tab:forecasts_imerg}
    \begin{tabular}{@{\extracolsep{5pt}} lccc}
      \toprule
     Inputs & 1-day & 3-day & 5-day \\ 
        \midrule
     Persistence & 1.1321 & 1.1497 & 1.1518 \\
    Climatology & \multicolumn{3}{c}{0.7696 (2000-2017)}  \\
    Climatology (weekly) & \multicolumn{3}{c}{\textbf{0.7687} (2000-2017)} \\
     \midrule
     SimSat & 0.8166 & 0.8201 & 0.8198 \\ 
     ERA & 0.8182 & 0.8224 & 0.8215 \\
      SimSat + ERA & \textbf{0.8134} & \textbf{0.8185} & \textbf{0.8185} \\
      \midrule
     ERA (2000-2017) & 0.8085 & 0.8194 & 0.8214 \\
      \bottomrule
    \end{tabular}
  \end{subtable}
\end{table}

\begin{figure*}[h]
\centering
\includegraphics[width=16.5cm]{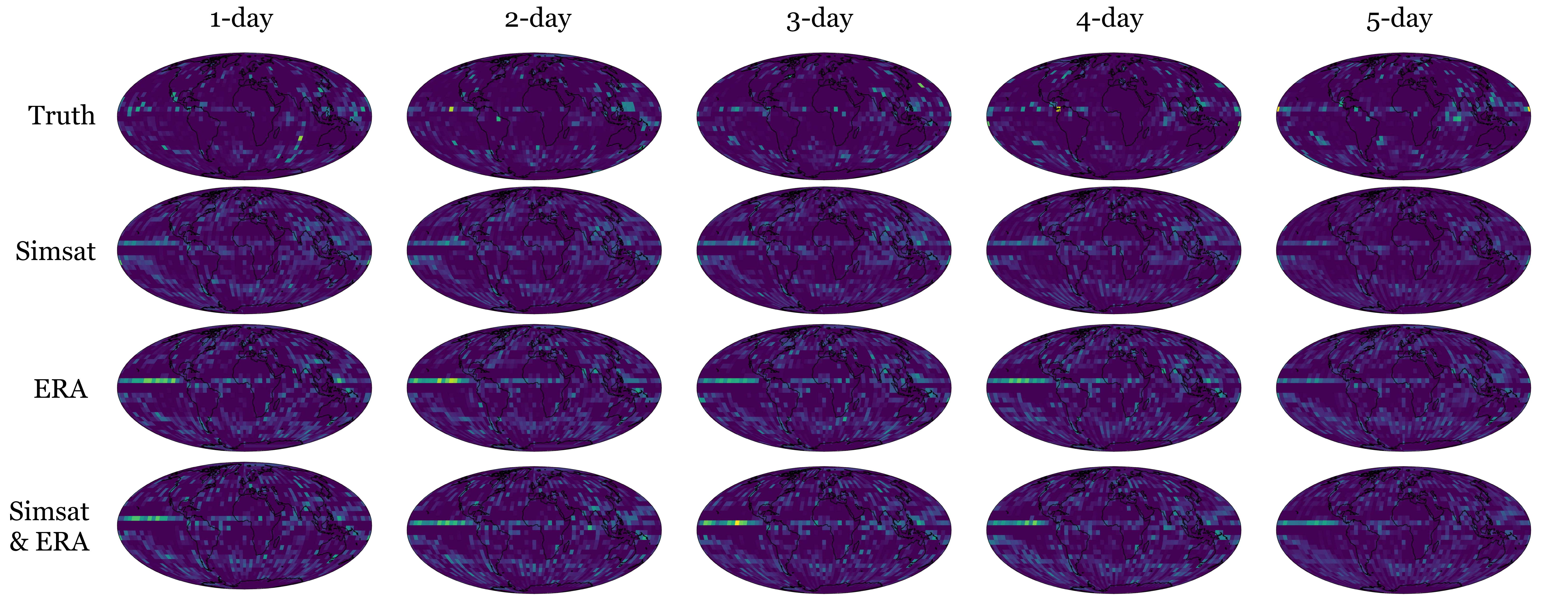}
\caption{ERA5 Precipitation forecasts on one random sample.}
\label{fig:lstm_forecasts}
\end{figure*}

\autoref{tab:forecasts} compares the neural model forecasts in different data settings when predicting precipitation from ERA5 and IMERG. Using the ERA5 precipitation as target, \autoref{tab:forecasts_era} shows that training from SimSat alone gives the worst results across the data settings. This confirms the difficulty in precipitation forecast from satellite data alone, which does not contain as much information about the atmospheric state as sophisticated reanalysis data such as ERA5. Importantly, the complementary benefits of utilizing data from both sources is already visible despite our simple concatenation setup, as training from both SimSat and ERA5 achieves the best results across all lead times (when holding the number of training instances constant). 


\autoref{fig:lstm_forecasts} shows example forecasts from one random input sequence across the different data settings for predicting ERA5 precipitation. We observe that the forecasts can capture the general precipitation distribution across the globe, but there is various degrees of blurriness in the outputs. As we shall discuss later in the paper, considering probabilistic forecasts would be a promising solution to blurriness, which might have arisen as the mean predicted outcome. 

We also see the importance in using a large training dataset, since extending the considered training instances to the full ERA5 dataset outperforms the baselines further in the 1-day forecasting regime (shown in the last rows).

\autoref{tab:forecasts_imerg} shows the forecast results when predicting IMERG precipitation. As before, the neural model's forecasting skill based on both SimSat and ERA input outperforms the other input settings. The higher observed RMSEs suggest that this is a considerably more difficult task, which we believe to be closely tied to IMERG featuring more extreme precipitation events (\autoref{fig:hist_rain}). In the next section, we investigate this issue further by considering a same-timestep precipitation estimation task. 


 A key limitation in our current experimental setup is that it requires all of ERA5, IMERG and SimSat channels to be available at each time step, limiting the range of our training data to April 2016 and onward. Nevertheless, our neural models significantly outperform persistence baselines. The fact that local climatology trained over longer time periods significantly outperforms our network model baselines suggests the development of alternative modelling setups that can make use of the full available datasets from each source.

\subsection{Same-Timestep Precipitation Estimation}
\label{sec:est}

We now describe a set of experiments for same-timestep precipitation estimation on IMERG. This analysis is done independently from the precipitation forecasting benchmark tasks, in order to provide an in-depth understanding of the challenges in modelling extreme precipitation events. 


We use a gradient boosting decision tree learning algorithm \cite[LightGBM]{ke2017lightgbm} in order to estimate same-timestep IMERG precipitation directly from ERA5 and SimSat. Our training set consists of $1$ million randomly sampled grid points/pixels within the time interval April $2016$ to December $2019$. We compare the (not latitude-adjusted) RMSE for two pixel sampling variants: A) unbalanced sampling, meaning grid points are chosen randomly from the raw data distribution and B) balanced sampling, in which we bin IMERG precipitation into the four classes defined in \autoref{fig:hist_rain} and sample grid points such that we end up with an equal amount of pixels per bin.

In \autoref{tab:precipestimation}, we find that taking a balanced sampling approach reduces the per-class validation RMSE of moderate, heavy and violent precipitation. This balanced sampling approach also has detrimental effects on the mean forecasting performance but not the macro-mean performance, as the `slight' class dominates the dataset and is misclassified more often. However, balancing the training set does result in a lower macro RMSE.

Designing an appropriate class-balanced sampling may play a crucial role toward improving predictions of extreme precipitation events. It is not quite clear how a per-pixel sampling scheme may be translated into a global output context approach such as in MetNet \cite{metnet} where each individual pixel's input distribution should be kept balanced, while training as many pixels per input data sample as possible for efficiency. A possible way of navigating this challenge would be to sample greedily, i.e. based on the currently most imbalanced pixel and combine this with learning rate adjustments for other pixels trained on the same frame based on how imbalanced these pixels are at that timestep.




\begin{table}[t]
\setlength{\tabcolsep}{4.5pt}
\begin{center}
\caption{Comparing RMSE Results with and without a class-balanced training dataset. The modelling task is same-timestep estimation of IMERG precipitation.} 
\label{tab:precipestimation}
\begin{tabular}{lccccccc}
      \toprule
     &  L & M & H & V & Mean & Macro \\
     \midrule
Unbalanced & & & & & \\    
    \midrule
ERA & \textbf{0.20} & 4.08 & 16.2 & 63.1 & 0.65 & 20.9 \\
SimSat & \textbf{0.20} & 4.38 & 16.8 & 54.1 & 0.65 & 18.9 \\
SimSat + ERA & \textbf{0.20} & 4.03 &  16.5 & 53.0 & \textbf{0.65} & 18.4 \\
     \midrule
Balanced & & & & & \\  
    \midrule
ERA & 1.05 & \textbf{2.75} & 12.4 & 58.0 & 1.40 & 18.6 \\
SimSat & 1.17 & 3.10 &13.3 &50.1 & 1.26 & 16.9 \\
SimSat + ERA & 1.30 & 3.15 & \textbf{11.8} & \textbf{44.3} & 1.38 & \textbf{15.1}\\
\bottomrule
\end{tabular}
\end{center}
\end{table}

\section{Discussion} \label{sec:discussion}
We outline the key challenges in global precipitation forecasting, our proposed solutions, we also discuss promising research avenues that can build on our work.

\subsection{Challenges} 
From our experiments, we identified a number of challenges inherent to data-driven extreme precipitation forecasting. 

\paragraph{Class imbalance} Extreme precipitation events, by their nature, rarely occur (see \autoref{fig:hist_rain}). In the context of supervised learning, this manifests as a class imbalance problem, in which a model might rarely predict extreme values. Designing an appropriate class sampling strategy (e.g. inverse frequency sampling) can mitigate this imbalance, as shown in our same-timestep prediction experiments.
Further, we believe that a mixture of pixelwise-weighting and balanced sampling could be a potential solution.



\paragraph{Probabilistic forecasts.} The current machine learning setup produces deterministic predictions, which may lead to an averaging of possible futures into a single blurry prediction. This limitation may be overcome with probabilistic modelling, which may take different forms. For instance, \citeauthor{metnet} made use of a cross-entropy loss over a categorical distribution to handle probabilistic forecasts. Stochastic video prediction techniques \cite{babaeizadeh_stochastic_2018} and conditional generative adversarial learning \cite{mirza_conditional_2014} have also been shown to produce realistic predictions in other fields.
Other relevant techniques that predict distribution parameters are Variational Auto-Encoders \cite{kingma_auto-encoding_2014} and normalizing flows \cite{rezende_variational_2016}.


\paragraph{Data normalisation.} Feature scaling is a common data-processing step for training machine learning models and well-understood to be advantageous \cite{bhanja_impact_2019}. Our current approach normalizes each variable using its global mean and standard deviation; This disregards any local spatial differences, which is important for modelling local weather patterns \cite{weyn_can_2019}. Previous work suggested that patch-wise normalisation may be appropriate \cite[Local Area-wise Standardization (LAS)]{gronquist_deep_2020}. We suggest studying a refinement to LAS, which adjusts the kernel size with latitude such that the spatial normalisation context remains constant (\textit{Latitude-Adjusted LAS}) per-channel image-size normalisation.

\paragraph{Data topology.} Lastly, the spherical input and output data topology of global forecasting contexts poses interesting questions to neural network architecture. While a multitude of approaches to handle spherical input topologies have been suggested, see \cite{llorens_jover_geometric_2020} for an overview, it seems yet unclear which approach works best. Our dataset might constitute a valuable benchmark for such research.


\subsection{Future research avenues}
Apart from overcoming the challenges outlined above, we have identified a variety of opportunities for further research.

\paragraph{Physics-informed multi-task learning. } Apart from using reanalysis data for model training, we do not currently exploit the fact that many aspects of weather forecasting are well-understood from a physical perspective. One way of informing model training of physical constraints would be to train precipitation forecasting concurrently with prediction of physical state variables, including temperature and specific humidity, in a multi-task setting, e.g. through using separate decoder heads for different variables (similarly to \citet{caruana_multitask_1997}). This approach promises to combine the advantages of data-driven learning with low-level feature regularisation through a physics-informed inductive bias. Multi-task learning can also be regarded as a form of data augmentation~\cite{shorten_survey_2019}, promising to further increase forecasting performance using real or simulated satellite data without requiring access to reanalysis data at inference time.

\paragraph{Increasing spatial resolution.} Data at higher spatial resolution tends to capture heavy and extreme precipitation events better but poses a number of challenges. Large sample batch sizes may lead to network activation storage that exceeds GPU global memory capacity even for distributed training. Apart from exploring TPU or nvlink-based solutions, another way would be to switch to mixed-precision or half-precision or employ techniques that trade-off memory for compute such as gradient checkpointing \cite{pinckaers_streaming_2019}. PyRain's dataloader efficiently maximises total disk throughput, which may itself become a bottleneck at very high resolutions. Storing all or part of the training data memmaps on one or several high-speed local SSDs may increase disk throughput a few-fold. 
Apart from memory and disk throughput, there is also a lack of suitably highly resolved historical climate data for pre-training \cite{rasp_weatherbench_2020}. One possible way of overcoming this would be to integrate high-resolution local forecasting model or sensor data into the training process \cite{franch_taasrad19_2020}, another exciting approach spearheaded in computational fluid dynamics \cite{jabarullah_khan_machine_2019} is to employ a multi-fidelity approach, where hierarchical variance-reduction techniques are employed to enable training to be performed at lower-resolution data as often as possible, thus minimising the need for training on high-resolution data.

\paragraph{Reducing IMERG Early Run lag time.} While the final IMERG product becomes available at a time lag of ca. 3-4 months, a preliminary, Early Run, product based on raw satellite data becomes available after ca. $4$ hours. We postulate that this lag could be further reduced if, instead of high-dimensional observational data, forecasting agencies were exchanging their locally processed low-dimensional embeddings derived from local encoder networks. Embeddings could then be feed into a late fusion network architecture similar to \citet[$\text{Multi}^3\text{Net}$]{rudner_multi3net_2019}. 

\paragraph{Multi-time-step loss function.} 
Numerical forecasting systems forward the physical state in time by following an iterative setting, where the output of the previous step is fed as input to the next step. As the update rules are identical for each step, it in principle suffices for neural networks to learn a single such update step and apply it multiple times during inference depending on the prediction lead time, thus reducing the number of trainable weights and potentially increase generalisation performance. To avoid instability issues inherent to iterative approaches \cite{rasp_weatherbench_2020}, model rollouts can be trained end-to-end \cite{mcgibbon_single-column_2019,brenowitz_prognostic_2018}.  \citet{weyn_improving_2020} pioneer this approach but limit themselves to just two time steps. To overcome device memory constraints in such a setting and to scale to a large number of time steps tollouts, iteration layers could be chosen to be reversible \cite{gomez_reversible_2017} such that activations can be computed on-the-fly during backpropagation and do not need to be stored in device memory.


\section{Conclusion}

We presented \textbf{RainBench}, a novel benchmark suite for data-driven extreme precipitation forecasting, and \textbf{PyRain}, an associated rapid experimentation framework with a fast dataloader. Both RainBench and PyRain are open source and well-documented. We furthermore present neural baselines for multi-day precipitation forecasting from both reanalysis and simulated satellite data. Despite our simple approach, we find that our neural baselines beat climatology and persistence baselines for up to $5$ day forecasts. In addition, we use a gradient boosting decision tree algorithm to study the impact of precipitation class balancing on regression in a precipitation estimation setting and present various forms of data exploration, including a correlation study.

In the near future, we will augment RainBench with real satellite data. We plan on also including historical climate data for pre-training. Concurrently, we will explore various directions for future research, as discussed above. In particular, we believe increasing the spatial resolution of our input data is crucial to closing the gap to operational forecasting models. Ultimately, we hope that our benchmark and framework will lower the barrier of entry for the global research community such that our work contributes to rapid progress in data-driven weather prediction, democratisation of access to adequate weather forecasts and, ultimately, help protect and improve livelihoods in a warming world.

\pagebreak

\clearpage
\section*{Acknowledgements}
This research was conducted at the Frontier Development Lab (FDL), Europe. The authors gratefully acknowledge support from the European Space Agency ESRIN Phi Lab, Trillium Technologies, NVIDIA Corporation, Google Cloud, and SCAN. The authors are thankful to Peter Dueben, Stephan Rasp, Julien Brajard and Bertrand Le Saux for useful suggestions.
\bibliography{references.bib}

\onecolumn

\section{Appendix}



\subsection{Pixel-wise precipitation class histograms (IMERG)}

We include pixel-wise precipitation class histograms derived from IMERG at native resolution ($0.1^\circ$) with max-pooling as downscaling to preserve pixel-wise extremes.

\begin{figure}[h]
\includegraphics[width=16cm]{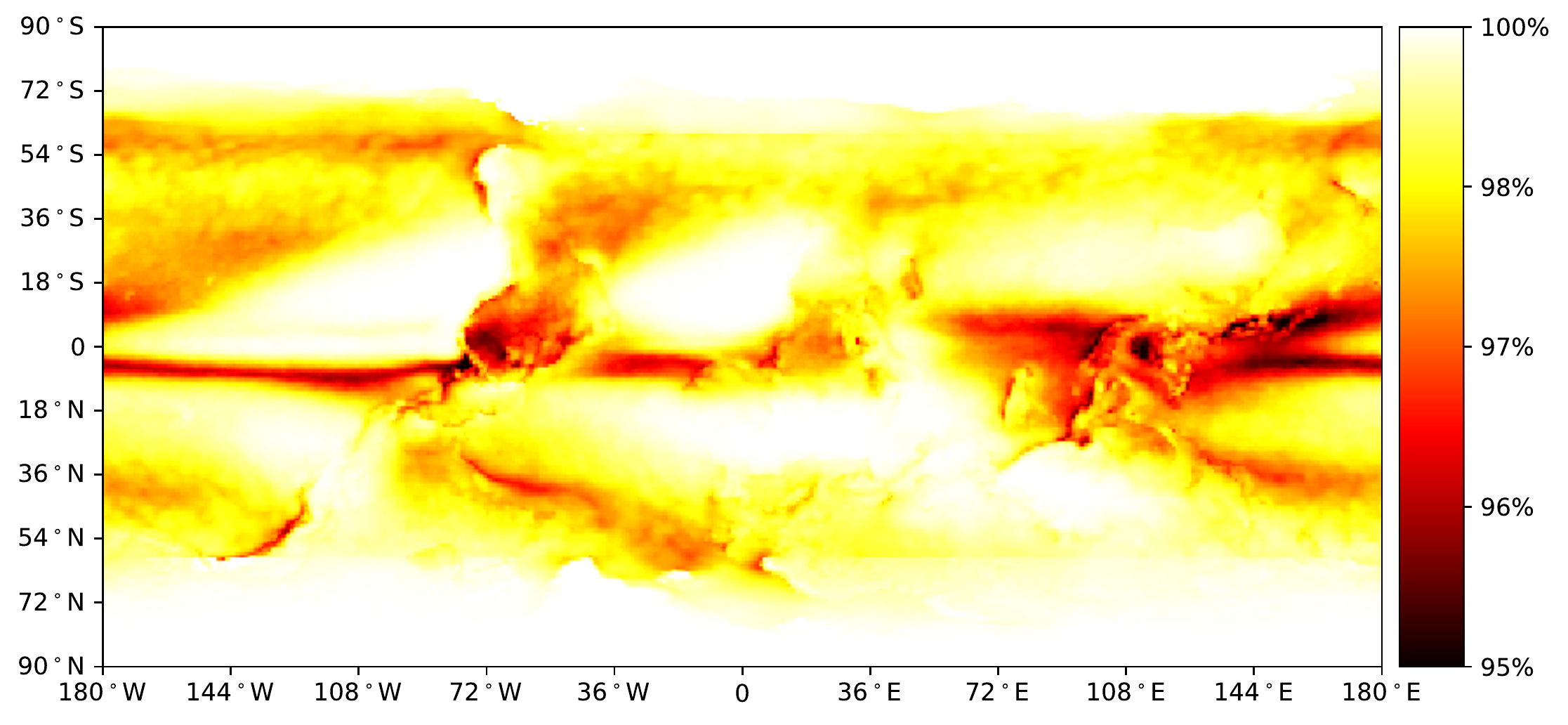}
\caption{Global distribution of slight rain events ($\%$ of total events)}
\end{figure}

\begin{figure}[h]
\includegraphics[width=16cm]{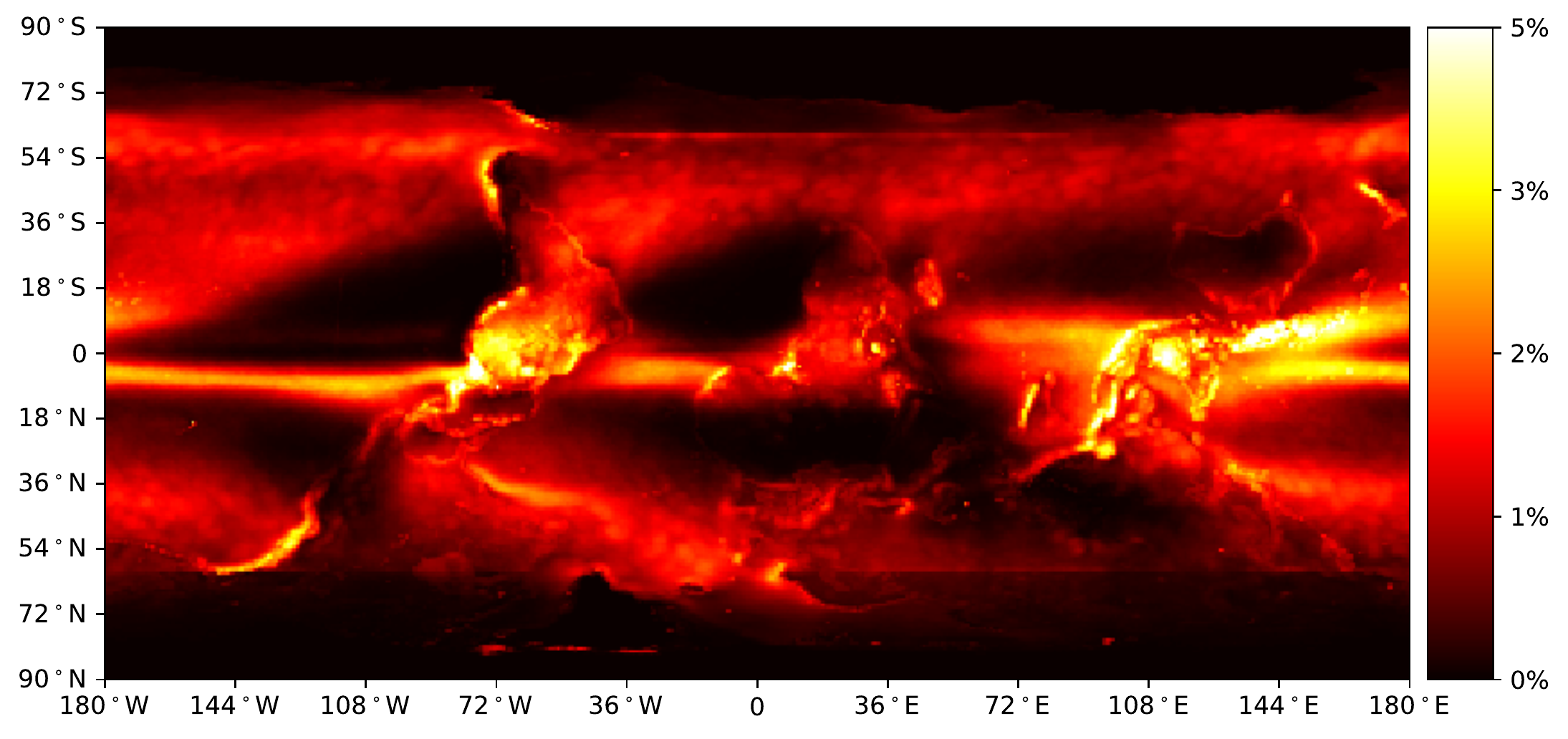}
\caption{Global distribution of moderate rain events ($\%$ of total events)}
\end{figure}

\begin{figure}[h]
\includegraphics[width=16cm]{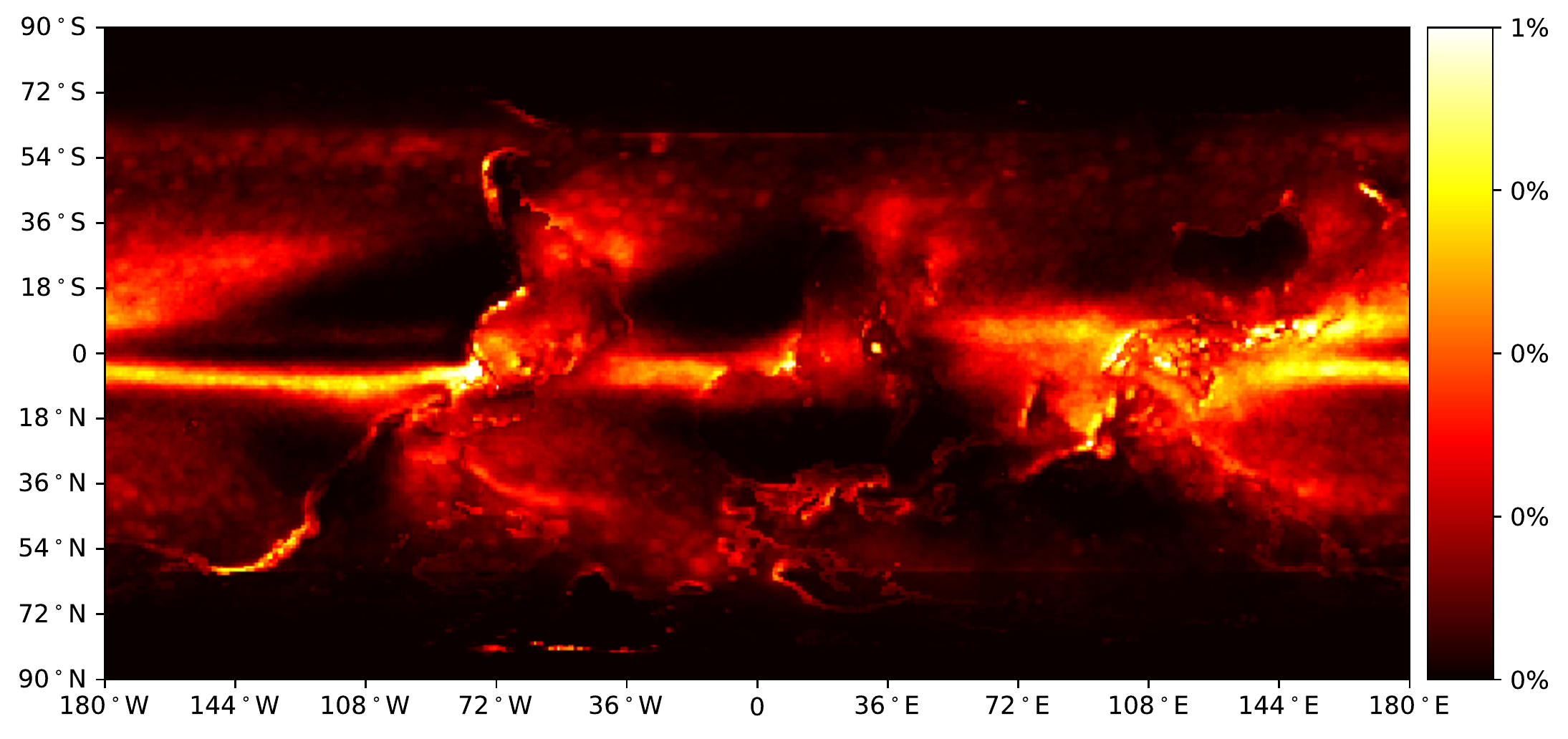}
\caption{Global distribution of heavy rain events ($\%$ of total events)}
\end{figure}

\begin{figure}[h]
\includegraphics[width=16cm]{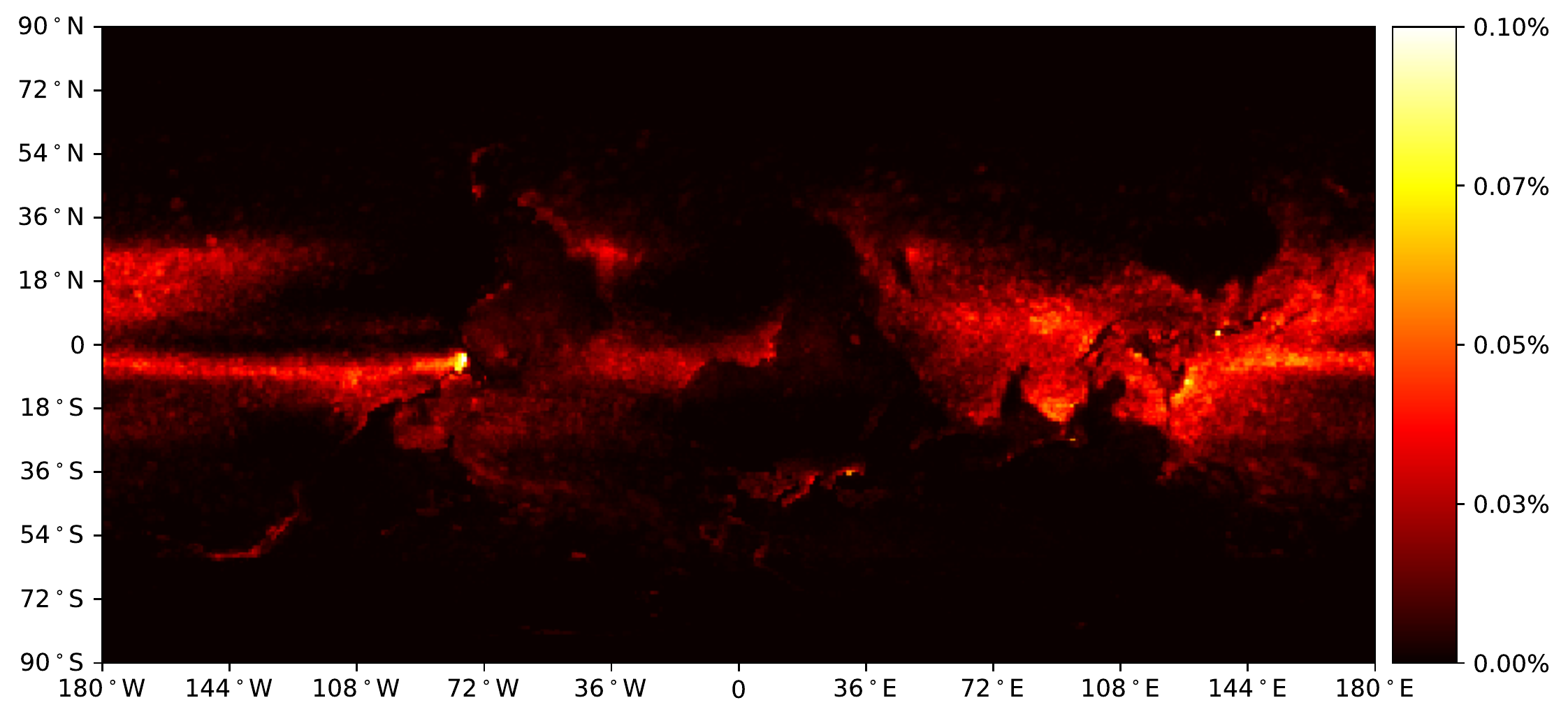}
\caption{Global distribution of violent rain events ($\%$ of total events)}
\end{figure}

\end{document}